# Machine Learning Approach to Remove Ion Interference Effect in Agricultural Nutrient Solutions

Byunghyun Ban[*]
Andong District Office
**Ministry of Employment and Labor**
Andong, Republic of Korea
bhban@kaist.ac.kr

Donghun Ryu
Machine Learning Team
**Imagination Garden Inc.**
Andong, Republic of Korea
dhryu@sangsang.farm

Minwoo Lee
Future Agriculture Team
**Imagination Garden Inc.**
Andong, Republic of Korea
hydrominus@sangsang.farm

*Abstract*— High concentration agricultural facilities such as vertical farms or plant factories consider hydroponic techniques as optimal solutions. Although closed-system dramatically reduces water consumption and pollution issues, it has ion-ratio related problem. As the root absorbs individual ions with different rate, ion rate in a nutrient solution should be adjusted periodically. But traditional method only considers pH and electrical conductivity to adjust the nutrient solution, leading to ion imbalance and accumulation of excessive salts. To avoid those problems, some researchers have proposed ion-balancing methods which measure and control each ion concentration. However, those approaches do not overcome the innate limitations of ISEs, especially ion interference effect. An anion sensor is affected by other anions, and the error grows larger in higher concentration solution. A machine learning approach to modify ISE data distorted by ion interference effect is proposed in this paper. As measurement of TDS value is relatively robust than any other signals, we applied TDS as key parameter to build a readjustment function to remove the artifact. Once a readjustment model is established, application on ISE data can be done in real time. Readjusted data with proposed model showed about 91.6 ~ 98.3% accuracies. This method will enable the fields to apply recent methods in feasible status.

I. INTRODUCTION

Development of IoT applications enabled the propagation of smart farms around agricultural industries. Even small farm owners now adopts smart farm facilities to reduce human resource. Highly concentrated facilities such as vertical farm or plant factories are one of the greatest issue in smart farm area. Aqua cultural methods such as NFT(nutrient film technique), DWT(deep water technique), water dripping or aeroponics are now considered as optimal solutions for high-concentrated smart farms because it becomes significantly efficient to control nutritional environment of whole facility with hydroponic approaches.

Opened hydroponic system discards once-used nutrient solution because the nutritional ions are already uptaken by plant roots. This approach tries to supply stable and balanced nutrition to plants. However, the wastage of nutrition water is hazard to environment so closed hydroponic systems becomes popular[1]. Closed system recycles the nutrient solution so it significantly reduces the cost and possibility of pollution.

It is one of the most important issue in closed system to maintain nutritional balance. Traditional approaches measure electrical conductivity and pH of the nutrient solution to monitor the fertilization[2-3]. However, those methods cannot distinguish the imbalance of nutrient ions and accumulation of useless components, caused by the difference between absorption rate of individual ions at plant roots. It requires additional maintenance labors such as foliar spray to avoid nutrition disorder which may lead to poor yield or death of crops.

Researchers recently suggests application of the ion selective electrodes(ISEs) approaches to measure individual ion's concentration from the nutrient solution [1, 4-6]. Readjustment of low-concentrated ions with ion selective measurement seems to enable avoidance of nutrient balance collapse. However, applying ISEs on nutrient solution measurement have significant problem: ion interference effect.

ISEs acquires concentration of ion by measuring electrical potential. ISE has ion-selective permeable film and the ion-exchange of specific ion fit for the film provides Nernst potential, described on equation (1). F is Faraday constant and z is the number of ions exchanged along the membrane.

$$E = \frac{RT}{zF} \ln \frac{[outside\ ion]}{[inside\ ion]} \quad (1)$$

However, because of the nature of ISE, ion-exchange along the membrane is interfered by other ions. This phenomenon is described with Nikolsky-Eisenman equation [7] given on equation (2) where $a$ is the activity, $k_i$ is the selectivity coefficient and $z_i$ is the exchange of interfering ion i along the membrane.

$$E = E_0 + \frac{RT}{zF} \ln \left[ a + \sum_i \left( k_i a_i^{\frac{z}{z_i}} \right) \right] \quad (2)$$

The activity parameter for target ion and the selectivity coefficients of all the other ions of solution are required to predict the magnitude of ion interference potential. It is not feasible. For example, Yamazaki's nutrient solution for lettuce [8] contains 23 different ions. Acquisition of those parameter with experiments to maintain agricultural fertilizer is not desirable. Liberti et al. suggested readjustment method using Gran's plots [9] but it also requires prior knowledge for some specific parameter.

We suggest a machine learning approach to train a readjustment function, which removes ion interference effect from ISE measured data without any prior knowledge chemicophysical parameter.

## II. ION INTERFERENCE EFFECT

We measured ion interference effect with ISEs to figure out the exact magnitude of the artifact. Also the data was used for the machine learning algorithm.

### A. Equipments for experiment

Ion selective electrodes were used to measure the concentration of ions. We used Vernier's Go Direct® sensor series. GDX-NO3 was used to measure $NO_3^-$ ion, and GDX-NH4 was used for $NH_4^+$, GDX-CA was used for $Ca^{2+}$, and GDX-K was used for $K^+$ ion concentration. The sensors were connected to Arduino to transform analog signal into digital value.

Yamazaki's nutrient solution for lettuce [8] was selected as experimental environment because it is one of the most popular fertilizer in hydroponics area. But we eliminated the low-concentrated ions in order to avoid ISE-unmeasurable ions. $KNO_3$, $Ca(NO_3)_2$-$4H_2O$ and $NH_4H_2PO_4$ were used to reproduce the simplified Yamazaki's solution. The experimental solution contains $K^+$, $NO_3^-$, $Ca^{2+}$, $NH_4^+$, which are measurable with ISEs, and $H_xPO_4^{3-x}$ families which are the only unmeasurable value. The concentration of $H^+$ is measurable with pH meter but we did not measured it because pH of the solution is around 7 so we did not expect any significant interference effect on pH meter.

### B. Calibration with machine learning algorithm

Vernier's manual recommended to measure 2 standard buffer solution to acquire 2 voltage value, and perform linear regression on exp(volt) term along the theoretical value of ion concentration. They presents voltage-concentration equation as equation (3), which fits for Nernst potential described on equation (1). Generally, α describes the sensitivity of sensor and β describes standard potential offset of the electrode.

$$\text{Concentration} = \alpha \exp(volt) + \beta \quad (3)$$

However, we figured out that Vernier's ISE does not work as equation (3). 10 times of buffer solution measurement showed that the correlation between concentrations and exp(volt) values showed exponential behaviors. The distribution of measured buffer solution data is described on Figure 1.

$R^2$ values on linear regression with Vernier's recommended method for $K^+$, $Ca^+$, $NO_3^-$, $NH_4^+$ were 0.9308, 0.9273, 0.9041, 0.9315. We figured out that Vernier's sensors actually work as equation (4).

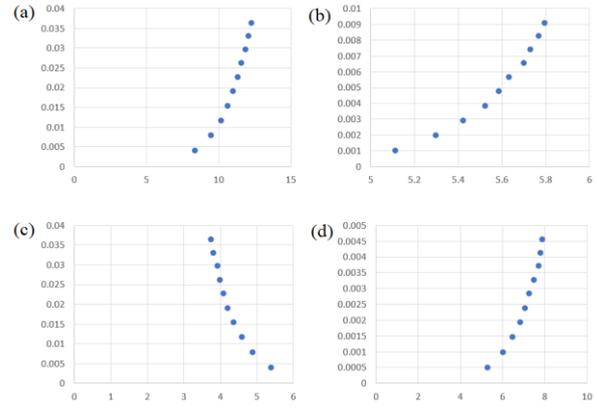

**Figure 1. Plot between exp(volt) and concentration.**
(a) $K^+$, (b) $Ca^{2+}$, (c) $NO_3^-$ and (d) $NH_4^+$

|  | 1 | 2 | 3 | 4 | 5 | 6 | 7 |
|---|---|---|---|---|---|---|---|
| $K^+$ | $\alpha = 3.31880e-5$, | | | $\beta = 0.57347$, | | $R^2 = 0.9957$ | |
| Concentration | 0.004 | 0.0078 | 0.0117 | 0.0154 | 0.0226 | 0.0296 | 0.0364 |
| ISE Voltage | 2.1268 | 2.2505 | 2.3197 | 2.3653 | 2.4282 | 2.4727 | 2.5067 |
| $Ca^{2+}$ | $\alpha = 9.16069e-11$, | | | $\beta = 3.17870$, | | $R^2 = 0.9831$ | |
| Concentration | 0.001 | 0.002 | 0.0029 | 0.0038 | 0.0057 | 0.0074 | 0.0091 |
| ISE Voltage | 1.6317 | 1.6674 | 1.6905 | 1.7089 | 1.729 | 1.7457 | 1.7573 |
| $NO_3^-$ | $\alpha = 5.452673$, | | | $\beta = -1.33850$, | | $R^2 = 0.9910$ | |
| Concentration | 0.004 | 0.0078 | 0.0117 | 0.0154 | 0.0226 | 0.0296 | 0.0364 |
| ISE Voltage | 1.6856 | 1.5864 | 1.5233 | 1.4753 | 1.4083 | 1.3656 | 1.321 |
| $NH_4^+$ | $\alpha = 6.457599e-6$, | | | $\beta = 0.82985$, | | $R^2 = 0.9910$ | |
| Concentration | 0.0005 | 0.001 | 0.0015 | 0.0019 | 0.0028 | 0.0037 | 0.0045 |
| ISE Voltage | 1.6672 | 1.7976 | 1.869 | 1.922 | 1.9829 | 2.0419 | 2.068 |

**Table 1. Calibration results**

$$\text{Concentration} = \alpha' \exp(\beta' \exp(volt)) \quad (4)$$

It means that Vernier's ISE does not work robustly. The behavior of sensor in thin solution differs from the behavior in dense solution. The concentration of buffer solutions, measured voltage and coefficients for equation (4) of each ion is presented in Table 1. Coefficients are driven with linear regression between log scaled concentration and exp(volt) values. The $R^2$ for each regression lines are also listed on Table 1. Regression models with equation (4) fits for the experimental data.

### C. Ion interference effect measurement

We prepared 100 times more concentrated version of simplified Yamazaki's solution. We added this ampoule 10 times to 1 liter water with 4 ISEs installed. The theoretical values of ionic concentration for each steps and the experimental values are presented on Table 2.

Comparison between the experimental values and the theoretical values are described on Figure 2. The interference effect becomes larger at higher-concentration status because the flux of undesirable ions through the membrane of ISE becomes greater magnitude of ion interference effect is understood as the distance between the blue and orange lines.

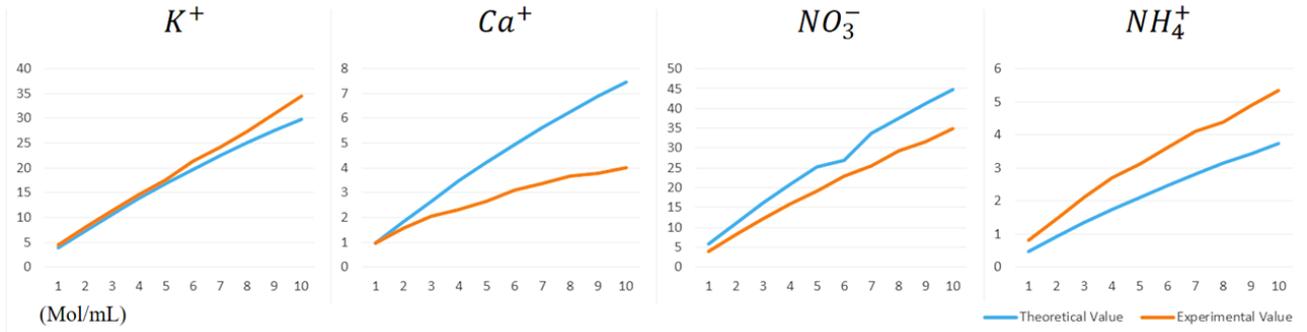

**Figure 2. Graph for ion interference effect** Theoretical values are described with blue lines and experimental values are in orange line.

| | 1 | 2 | 3 | 4 | 5 | 6 | 7 | 8 | 9 | 10 |
|---|---|---|---|---|---|---|---|---|---|---|
| **K** | | | | | | | | | | |
| Theoretical Value | 3.8835 | 7.327 | 10.689 | 13.87 | 16.885 | 19.747 | 22.467 | 25.055 | 27.521 | 29.873 |
| Experimental Value | 4.36955 | 7.96121 | 11.3368 | 14.5715 | 17.55282 | 21.32524 | 24.19406 | 27.28197 | 30.85615 | 34.4981 |
| **Ca** | | | | | | | | | | |
| Theoretical Value | 0.97087 | 1.832 | 2.672 | 3.467 | 4.221 | 4.937 | 5.617 | 6.264 | 6.88 | 7.468 |
| Experimental Value | 0.95579 | 1.54849 | 2.04874 | 2.31673 | 2.650174 | 3.119853 | 3.356937 | 3.65157 | 3.790148 | 4.00463 |
| **NO3** | | | | | | | | | | |
| Theoretical Value | 5.82524 | 10.991 | 16.033 | 20.804 | 25.327 | 26.92 | 33.7 | 37.582 | 41.281 | 44.81 |
| Experimental Value | 4.04472 | 8.24084 | 12.1232 | 16.0027 | 19.1123 | 22.95228 | 25.57974 | 29.36491 | 31.64328 | 34.8821 |
| **NH4** | | | | | | | | | | |
| Theoretical Value | 0.48544 | 0.916 | 1.336 | 1.734 | 2.111 | 2.468 | 2.808 | 3.132 | 3.44 | 3.734 |
| Experimental Value | 0.81084 | 1.46115 | 2.09361 | 2.68363 | 3.104724 | 3.616726 | 4.101231 | 4.388624 | 4.899455 | 5.34784 |

**Table 2. Ion interference effects**

### III. METHOD

A readjustment method to remove ion interference effect is proposed in this section. In order to provide a rapid and feasible solution to the industrial field, we avoided evaluation of chemicophysical parameters. We applied simple machine learning algorithm which doesn't even requires any GPU devices.

#### A. Equipments

The regression process was performed on Microsoft's Azure VM standard D4s v3, which has 4 vcpus and 16 GiB memory. The operation system is ubuntu 18.04 and we used R and R Studio Server for data analysis and regression.

#### B. Data preperation

Experimental data on Table 2 from ion interference effect measurement was used. We randomly separated 3 columns for test data and the other 7 columns for training data. Column 5, 7 and 9 were selected as test data and the others were fed to the model for training.

#### C. Readjustment method

We applied simple readjustment method in order to application of the trained model on readjustment as simple as possible because real-time and on-site readjustment is one of the biggest issue in the agricultural industry [1].

As TDS(total dissolved solids) is calculated regardless of ion composition in the solution, we took TDS value as key independent variable. The readjustment method is described as equation (5), where $C_r$ is readjusted value for concentration, $C_{ISE}$ is experimentally driven value for concentration and $\mu$ is a function of TDS value.

$$C_r = \mu(TDS) \times C_{ISE} \quad (5)$$

We assumed the concentration of each ion and the magnitude of ion interference effect as functions of TDS in Yamazaki's nutrient solution system because the imbalance of ionic balance does not occur fast. Also, with continuous readjustment on the nutrient solution, the ionic balance is maintained continuously.

### D. Regression of μ function

TDS values were calculated from the theoretical value of ion concentrations from Table 2. TDS is calculated as equation (6). $M_{ion}$ is the molecular weight and [ion] is molarity.

$$\sum M_{ion} \times [ion] \quad (6)$$

From the definition of μ, converged form of $\alpha$ function is defined as equation (7).

$$\mu(TDS) = \frac{C_r}{C_{ISE}} \quad (7)$$

And we expect the readjusted value $C_r$ to converge toward theoretical value $C_t$. Therefore, we launched regression to fit equation (9).

$$\mu(TDS) \approx \frac{C_t}{C_{ISE}} \quad (8)$$

The values of μ(TDS) driven from 7 points of training data are plotted on Figure 3 as blue dots.

As the behavior of voltage along ion concentration from equation (1) and the interference effect driven from equation (2) are both logarithm, we expected that the error correction coefficient μ to show logarithmic distribution. However, quadratic regression fit much better than log regression. Mean $R^2$ from log regression was 0.70615 while quadratic regression showed 0.96158. The regression lines are described as orange lines on Figure 3. Parameters for quadratic regression model equation (10) and $R^2$ values for the models are also presented in Table 3.

|  | A | B | C | $R^2$ |
|---|---|---|---|---|
| $K^+$ | -3e-08 | 1e-04 | 0.8322 | **0.9478** |
| $CA^2$ | -2e-08 | 4e-04 | 0.8286 | **0.9871** |
| $NO_3^-$ | 3e-08 | -2e-04 | 1.521 | **0.9255** |
| $NH_4^+$ | -3e-08 | 3e-04 | 0.9382 | **0.9859** |

**Table 3. Regression Results**

$$\mu(TDS) = A \times TDS^2 + B \times TDS + C \quad (10)$$

## IV. EXPERIMENT

### A. Evaluation on training data

To evaluate the trained model on training data, we calculated μ(TDS) from equation (10), rather than equation (9) because value from equation (9) will provide 100% correctness, which is meaningless. The driven μ are multiplied with $C_{ISE}$ to show whether the ion interference effect were removed. The results are presented on Figure 4. The model removed 45.8752%, 63.2386%, 81.5443% and 92.434% of ion interference effect from $K^+$, $Ca^{2+}$, $NO_3^-$ and $NH_4^+$. $R^2$ of $Ca^{2+}$ is greater than that of $NO_3^-$ but $NO_3^-$ showed better result.

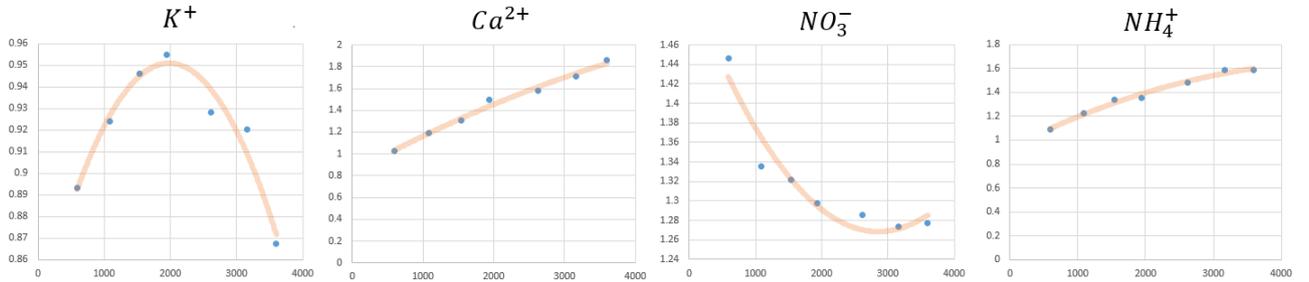

**Figure 3. μ(TDS) values with regression lines.** Horizontal axis are independent variable, TDS, while vertical axis is μ(TDS) driven from equation (9).

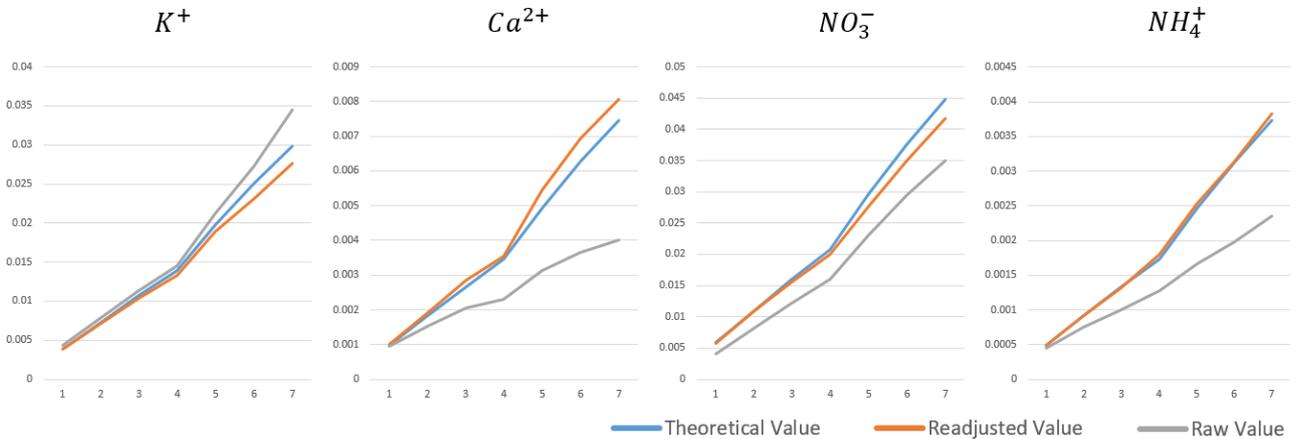

**Figure 4. Results of trained model on training data**

## B. Evaluation on test data

Evaluation on test data was performed by trained model. We apply untrained TDS value as input to μ function, and multiplied that to $C_{ISE}$ of test data.

## V. RESULTS

Test data was applied on the trained quadratic model. The result is described on Figure 5 and Table 3-4. The error was defined as equation (11), where C stands for both $C_{ISE}$ and corrected concentration. Accuracy was defined as 100% - error.

$$\text{Error} = \frac{C_t - C}{C_t} \times 100\% \quad (11)$$

The errors on whole date are presented on Figure 3. Except only one case, column 5 from $K^+$, proposed method successfully removed the ion interference effect. Especially, $NH_4^+$ model showed remarkable result. Average accuracy values are presented on Table 4. The model worked on both training data and test data. The accuracy between training data and test data from $Ca^+$ was significantly different. However, the proposed method successfully increased the accuracy of test data, 59.331%, into 93.209 %.

## VI. CONCLUSION

Machine learning algorithm with low computational complexity to remove ion interference effect was presented. Ion interference effect exists and gets stronger in denser environment. Quadratic regression on equation (9) successfully removed ion interference effect from almost all the data point, except only one case. As the computational resource for using pre-trained quadratic model is small enough, the presented method is highly feasible to industries. Even an embedded device can run the inference process in milliseconds. We expect applying higher-variance machine learning algorithms such as deep neural network, rather than high-bias and low variance model such as quadratic regression, would enhance the ability of artifact removal.

ACKNOWLEDGMENT

Special thanks to Janghun Lee for selecting ISE devices and establishing measurement pipeline with Arduino.

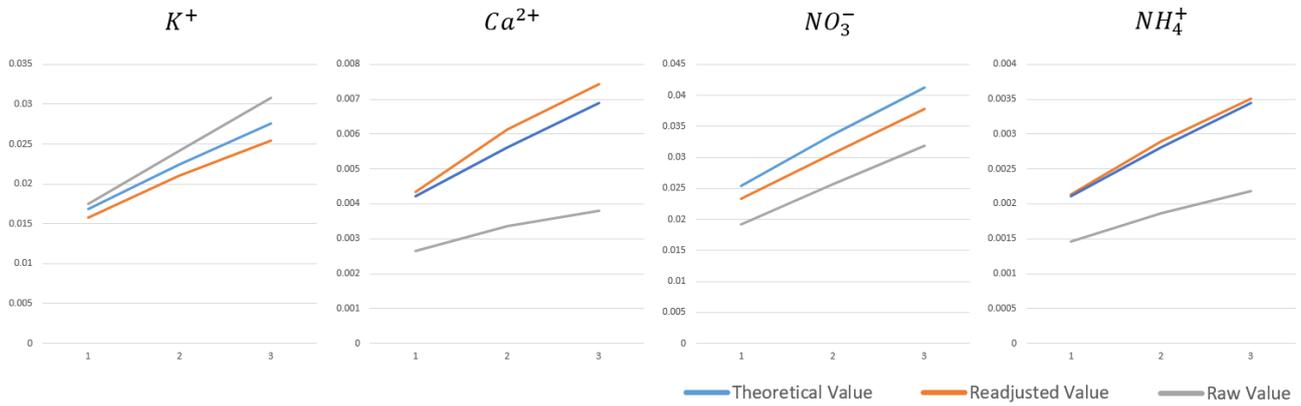

Figure 5. Result of trained model on test data

|  | Error rate from | 1 | 2 | 3 | 4 | 5 | 6 | 7 | 8 | 9 | 10 | Average |
|---|---|---|---|---|---|---|---|---|---|---|---|---|
| $K^+$ | Raw data | 11.946 | 8.2106 | 5.6993 | 4.7419 | 3.6743 | 7.7369 | 7.4555 | 8.6747 | 11.922 | 15.301 | 8.536235 |
|  | Proposed Method | 1.4019 | 2.0399 | 3.284 | 4.3368 | 6.3218 | 4.2945 | 6.5711 | 7.7969 | 7.5201 | 7.3183 | **5.088521** |
| $Ca^{2+}$ | Raw data | 2.4928 | 15.887 | 23.505 | 33.26 | 37.207 | 36.693 | 40.086 | 41.514 | 44.714 | 46.157 | 32.15142 |
|  | Proposed Method | 3.1302 | 4.1258 | 6.7386 | 2.0261 | 3.1154 | 10.093 | 9.1661 | 10.744 | 8.0917 | 8.1425 | **6.537297** |
| $NO_3^-$ | Raw data | 30.875 | 25.106 | 24.333 | 22.927 | 24.325 | 22.225 | 23.776 | 21.486 | 22.948 | 21.714 | 23.97158 |
|  | Proposed Method | 2.2965 | 0.3383 | 2.7885 | 3.9622 | 7.689 | 6.4383 | 9.054 | 6.6825 | 8.4805 | 6.8599 | **5.458977** |
| $NH_4^+$ | Raw data | 8.193 | 18.207 | 25.298 | 26.132 | 30.999 | 32.497 | 33.712 | 36.902 | 36.697 | 36.999 | 28.56361 |
|  | Proposed Method | 1.4263 | 0.4125 | 0.7996 | 3.9279 | 1.3726 | 2.478 | 3.1858 | 0.1296 | 1.9538 | 2.634 | **1.832009** |

**Table 4. Error rate of raw data and proposed method (in %)** The column 5, 7 and 9 are training data and the other columns are used for training.

|  | $K^+$ | | $Ca^{2+}$ | | $NO_3^-$ | | $NH_4^+$ | |
|---|---|---|---|---|---|---|---|---|
|  | Training | Test | Training | Test | Training | Test | Training | Test |
| Raw data | 91.0985 | 92.3161 | 71.499 | 59.331 | 75.9049 | 76.3167 | 73.6818 | 66.1971 |
| Proposed method | **95.6468** | **93.1957** | **93.5715** | **93.2089** | **95.8048** | **91.5922** | **98.3132** | **97.8293** |

**Table 5. Accuracy of raw data and proposed method (in %)**

## References

[1] Cho W J, Kim H J, Jung D H, Kim D W, Ahn T I, Son J E. On-site ion monitoring system for precision hydroponic nutrient management. Comput Electron AGR. 2018; 146: 51-58.
https://doi.org/10.1016/j.compag.2018.01.019.

[2] Bamsey M, Graham T, Thompson C, Berinstain A, Scott A, Dixon M. Ion-specific nutrient management in closed systems: the necessity for ion-selective sensors in terrestrial and space-based agriculture and water management systmes. Sensors. 2012; 12(10): 13349-13392.
https://doi.org/10.3390/s121013349.

[3] Katsoulas N, Savvas D, Kitta E, Bartzanas T, Kittas C. Extension and evaluation of a model for automatic drainage solution management in tomato crops grown in semi-closed hydroponic systems. Comput Electron AGR. 2015; 113: 61-71.
https://doi.org/10.1016/j.compag.2015.01.014.

[4] Kim H J, Son D W, Kwon S G, Roh M Y, Kang C I, Jung H S. PVC membrane-based portable ion analyzer for hydroponic and water monitoring. Comput Electron Agric. 2017; 140: 374-385.
https://doi.org/10.1016/j.compag.2017.06.015.

[5] Ryan S Knight, Mark Lefsrud. Automated Nutrient Sensing and Recycling. 2017 ASABE Annual International Meeting. 2017; 1701609. https://doi.org/10.13031/aim.201701609.

[6] Jung D H, Kim H J, Kim W K, Kang C I, Choi G L. Automated Sensing and Control of Hydroponic Macronutrients Using a Computer-controlled System. ASABE. 2013; 131594020.
https://doi.org/10.13031/aim.20131594020.

[7] Hall, Denver G. "Ion-selective membrane electrodes: a general limiting treatment of interference effects." *The Journal of Physical Chemistry* 100.17 (1996): 7230-7236.
https://doi.org/10.1021/jp9603039

[8] Yamazaki K. Nutrient solution culture. Tokyo, Japan: Pak-kyo Co; 1982. Japanese

[9] Liberti, Arnaldo, and Marco Mascini. "Anion determination with ion selective electrodes using Gran's plots. Application to fluoride." Analytical Chemistry 41.4 (1969): 676-679.